\documentclass[letterpaper, 10 pt, conference]{ieeeconf}  % Comment this line out if you need a4paper

\IEEEoverridecommandlockouts                              % This command is only needed if 
\usepackage{xcolor}
\usepackage{colortbl}
\definecolor{blue}{HTML}{d3cee7}
\usepackage[T1]{fontenc}
 \usepackage{graphicx} 
 \usepackage{amssymb}
 \usepackage{amsmath}
 \usepackage{booktabs}
 \usepackage{multirow}
 \usepackage{tabularx}
 
 \usepackage{url}
 \usepackage{xurl}     
\usepackage{hyperref} 
\usepackage{cleveref}

\title{\LARGE \bf
PlantShade: Predicting Plant Shadows for Lighting-Aware Robotic Agricultural Operation
}

\author{ 
Longchao Da$^{1,*}$, 
Xiaoou Liu$^{1,*}$, 
Xingjian Li$^{2,*}$, 
Lirong Xiang$^{2,\dagger}$, 
and Hua Wei$^{1,\dagger}$% 
\thanks{$^{*}$Longchao Da, Xiaoou Liu, and Xingjian Li contributed equally.}% 
\thanks{$^{\dagger}$Lirong Xiang and Hua Wei are co-corresponding authors ({\tt\small lxiang@cornell.edu}, {\tt\small hua.wei@asu.edu}).}% 
\thanks{$^{1}$Longchao Da, Xiaoou Liu, and Hua Wei are with the School of Computing and Augmented Intelligence, Arizona State University.}% 
\thanks{$^{2}$Xingjian Li and Lirong Xiang are with the Automation and Robotics Laboratory, Department of Biological and Environmental Engineering, Cornell University.}% 
\thanks{The work was partially supported by NSF award \#2442477, \#2550203 and \#2536297. The views and conclusions in this paper should not be interpreted as representing any funding agencies.}
}

\begin{document}

\maketitle
\thispagestyle{empty}
\pagestyle{empty}

%%%%%%%%%%%%%%%%%%%%%%%%%%%%%%%%%%%%%%%%%%%%%%%%%%%%%%%%%%%%%%%%%%%%%%%%%%%%%%%%
\begin{abstract}

Plant growth and agricultural production form the foundation of a country’s sustainable development and directly impact human livelihoods. Recent advances in frontier artificial intelligence have enabled scientific agriculture with strong potential to improve crop productivity. In this paper, we identify the importance and inherent complexity of plant shade simulation, as shading is a critical factor influencing plant growth. To advance this field and promote broader societal benefits, we focus on two main contributions. First, we introduce a comprehensive plant growth and shade dataset covering four plant species, including soybean, tomato, sugarbeet, and strawberry. The dataset includes top-down viewpoints with a supplementary light along a circular trajectory, casting dynamic shadows across multiple growth stages and diverse observation complexities. Second, we propose generative shade simulation based on diffusion models, enabling realistic shade generation for unseen plants and supporting downstream robotic tasks such as perception, lighting control, and view planning. The model incorporates temporal conditioning to facilitate flexible shade simulation across different time stages. We conduct both quantitative and qualitative evaluations to assess model performance. This work provides a foundational study for plant-aware shade modeling and has meaningful implications for broader agricultural and robotic applications.

\end{abstract}

%%%%%%%%%%%%%%%%%%%%%%%%%%%%%%%%%%%%%%%%%%%%%%%%%%%%%%%%%%%%%%%%%%%%%%%%%%%%%%%%
\section{INTRODUCTION}

\begin{figure}[t!]
    \centering
    \includegraphics[width=1\linewidth]{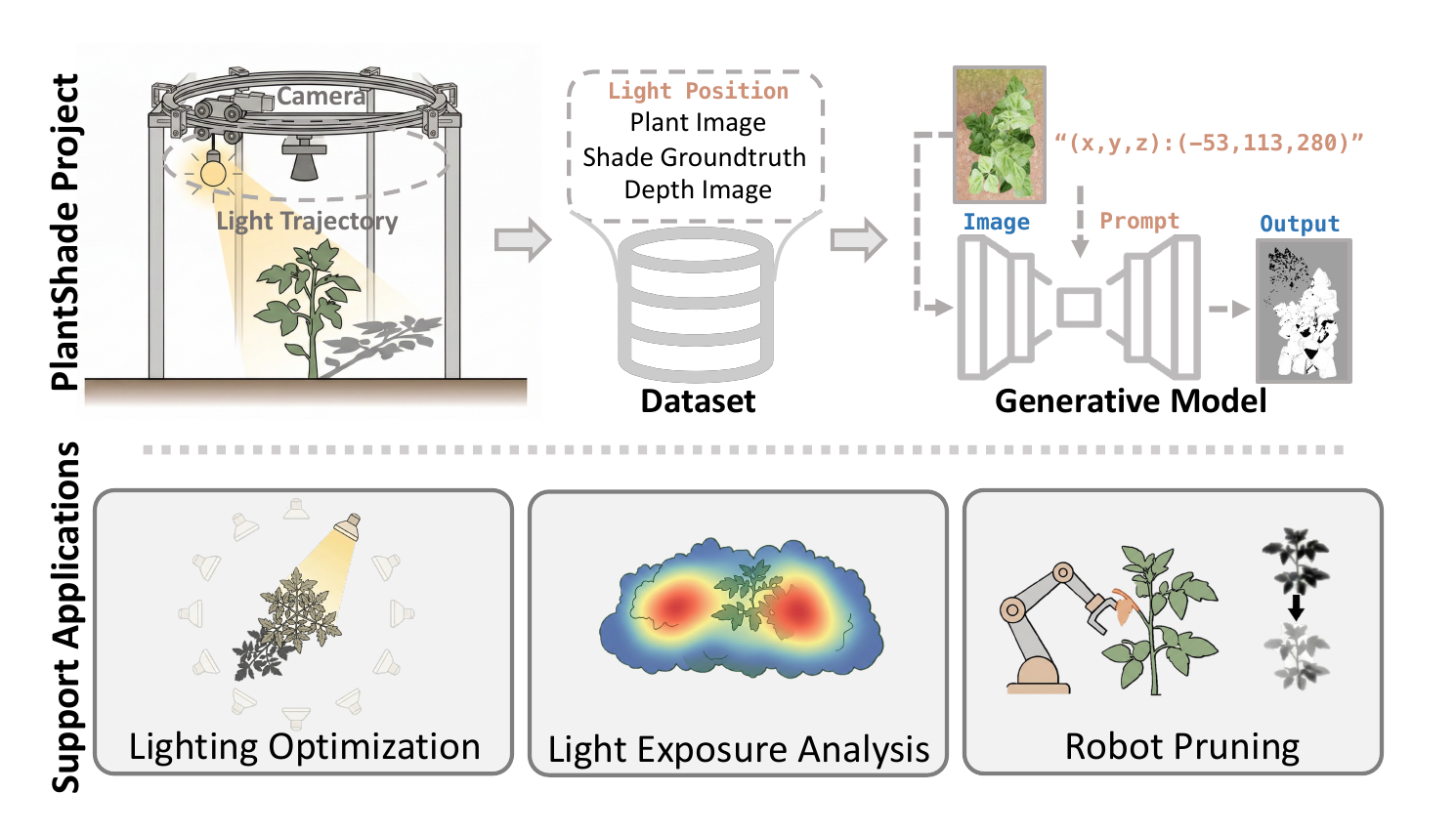}
    \vspace{-6mm}
    \caption{\textbf{Top:} The PlantShade system captures plant images under a controlled robotic lighting setup, where a movable supplemental light follows a circular trajectory to generate shadow variations. The dataset is used to train a light-conditioned generative model for shadow prediction. Given a plant image and light-position prompts, the model produces a shadow map. \textbf{Bottom:} The predicted shadow enables applications including lighting optimization through placement-adjust, light exposure analysis via heatmap estimation, and shade-aware robotic pruning by simulating branch removal and assessing impacts on self and mutual-shading.}
    \label{fig:intro}
    \vspace{-5mm}
\end{figure}

% \paragraph{What is the problem?}
Agricultural robots rely on vision and simulation for perception, planning, and control in crop environments. A central source of variability is \emph{shading}: the pattern of light and shadow over the canopy changes with sun angle, supplemental lighting, and plant geometry. We aim to simulate how plant shadows appear and evolve under different lighting conditions from a fixed camera view (e.g., a nadir camera on a mobile or gantry system in Fig.~\ref{fig:intro}). This enables robots to reason about illumination and ensures that synthetic training data and digital twins capture realistic shade dynamics. Shading affects both visual perception and underlying plant physiology \cite{sukhova2021mathematical}, making accurate modeling essential for lighting-aware planning and sim-to-real transfer.

% \paragraph{Why is it interesting and important?}
Shading also directly influences photosynthesis and yield by adjusting the daylight integral received by the crop~\cite{zamani_greenhouse_light}, and it dominates the visual appearance of crops in the field. The ability to generate realistic, time-varying plant shadows supports several robotic and agricultural goals: optimizing supplemental lighting without costly physical trials~\cite{WACKER2022129}, generating large-scale synthetic data for perception under diverse lighting~\cite{baker2024scalable}, and evaluating lighting strategies in simulation before deployment~\cite{shade_economics_sim}. These dynamics matter for agricultural robots, where shadow-aware modeling supports perception and sim-to-real transfer, growth monitoring, and view planning. To our knowledge, no prior work provides both a dedicated image dataset and a generative model for plant-specific shade under controllable lighting, leaving a gap for robot-oriented applications.
% \paragraph{Why is it hard?}

Plant shade prediction is challenging for three reasons. First, shadows are \emph{growth-dependent}: the same light source produces different shadows as the plant develops from seedling to mature canopy, so any pipeline must account for temporal change in structure~\cite{XIAO20230082}. Second, they are \emph{view- and light-dependent}: a fixed sensor sees shadows that evolve as the sun or supplemental lights move, requiring joint representation of geometry and light configuration~\cite{retkute_light_dynamics}. Third, plant canopies are semi-transparent and finely structured, yielding soft, complex shadow boundaries~\cite{shade_accuracy_bailey}. Naive approaches fail: using a single static shadow or simple geometric casting ignores growth and light variation; simple geometric projections ignore the path of moving light sources; and traditional computer vision models trained on rigid objects struggle to generalize to the layered plant geometry.

% \paragraph{Why hasn't it been solved before?}
This gap between structural complexity and predictive requirements is not sufficiently addressed by existing modeling frameworks. On the one hand, biophysical plant models focus on biomass and physiology~\cite{bailey2019helios,groimp,zhou2020cplantbox}; when they compute light interception, rendering is typically offline and not designed for real-time, camera-view imagery for robot perception or training. On the other hand, learning-based shadow generation methods target generic scenes with rigid objects and static lighting and do not handle the temporal and morphological variability of living plants. Existing agricultural shadow datasets either use static scenes or lack multi-species, multi-stage coverage with controllable light trajectories. Thus, there has been no pipeline that combines real-time plant shade data generation with a learned model that generalizes to unseen structure and growth stages.

% \paragraph{How do we solve it?}
We introduce \textbf{PlantShade}\footnote{\mbox{\url{https://darl-genai.github.io/PlantShade/}}: code and project page.}, the first dataset and real-time simulation framework and dataset for plant shade dynamics and robotic agricultural operations. %To overcome the lack of data and the need for growth- and light-variation, 
We build a dataset of over 38,500 image pairs from a Helios--UE5 pipeline, with four crop species, multiple growth stages (7--119 days), and single- and multi-plant layouts, with a supplementary light moving along a circular trajectory to produce dynamic shadow ground truth from a nadir view. To enable realistic shade generation without full 3D rendering, we propose a ControlNet-conditioned diffusion model that takes a reference plant image and a textual description of the light position and outputs the corresponding shadow map, supporting unseen plants and time steps. We then verify the approach with quantitative metrics %(MSE, mIoU, B-IoU, LPIPS) 
and qualitative analysis across species and layout complexity.

% \paragraph{Contributions.}
Our main contributions are:
\begin{itemize}
    \item A large-scale \textbf{plant shade dataset} (38,500 pairs) with four plant species, multiple growth stages, and single/multi-plant layouts, with circular multi-view shadow capture from a top-down view suitable for robot-mounted cameras.
    \item A \textbf{generative shade simulation model} based on ControlNet-conditioned diffusion that, given plant appearance and light position, generates realistic shadow maps for unseen plants and time steps, enabling rapid lighting scenario evaluation.
    \item \textbf{Extensive quantitative and qualitative experiments} across plant types and layout complexity, demonstrating feasibility and identifying directions for improvement for downstream robotic tasks.
\end{itemize}

\section{Related Work }

\subsection{Plant growth modeling and simulation}

Modeling and simulating plant growth is fundamental to understanding agricultural production and optimizing crop management. Classical approaches can be broadly categorized into process-based models and functional-structural plant models (FSPMs). Process-based models such as DSSAT~\cite{dssat}, APSIM~\cite{holzworth2018apsim}, and ALMANAC~\cite{xie2003almanac} simulate physiological processes at the canopy or field level, predicting biomass and yield based on environmental inputs including location, temperature, and water availability. However, PBMs typically abstract away the three-dimensional architecture of plants, making it difficult to reason about spatially varying factors such as shading.

FSPMs address this limitation by explicitly representing plant structure at the organ level. The L-system formalism~\cite{zamir200l-system} encodes recursive branching rules that generate realistic plant architectures, while models such as GreenLab~\cite{wang2024functional} couple structural development with source-sink dynamics for biomass allocation. Platforms like CPlantBox~\cite{zhou2020cplantbox} and OpenAlea~\cite{pradal2015openalea} provide modular environments for simulating structural growth, light interception, and photosynthesis. While FSPMs enable fine-grained light-interception and detailed structural analysis, the rendered outputs are typically low-fidelity rather than photorealistic. This limits their applicability in perception-driven tasks, where high-fidelity images are essential for computer vision systems. 
%While these models capture growth dynamics with physical plausibility, they rely on manually designed rules and parameters, limiting their generalizability across species and environmental conditions.

%However, existing plant growth models and datasets largely neglect the role of shading which is a factor that critically influences photosynthesis, canopy microclimate, and ultimately crop yield. Most datasets capture plants under controlled or diffuse lighting without modeling the dynamic shadows cast by the plants themselves or by changing light conditions. Our work addresses this gap by introducing a multi-species plant growth dataset that explicitly captures shade dynamics under a controlled circular light trajectory across multiple growth stages, providing a resource that bridges plant development and shade interaction.

\subsection{ Shadow modeling and generation}

% Shadows encode rich information about scene geometry, light source properties, and spatial relationships. Traditional shadow rendering relies on physics-based approaches such as shadow mapping, shadow volumes, and ray tracing~\cite{williams1978casting,hu2024unveiling}, which require explicit 3D scene geometry and lighting parameters. In agricultural contexts, Huang et al.~\cite{huang2024synthetic} generated a synthetic shadow dataset of 50,000 photorealistic images with physics-based shadow masks for agricultural field settings using 3D modeling software. While the dataset is physically accurate and large-scale, it is limited to a resolution of 512$\times$512 and includes only a few plant models.

% Learning-based shadow generation has progressed from GAN-based methods~\cite{liu2020arshadowgan, zhang2019shadowgan} to diffusion-based approaches~\cite{liu2024shadowdiffusion}. However, existing shadow generation methods are predominantly designed for generic scene compositing, which involves inserting rigid objects into background images. They do not account for the unique challenges of plant shade simulation, where complex canopy structures with intricate self-shadowing patterns change dramatically as the plant grows and as the light source moves. Our approach directly addresses these challenges by proposing a generative shade simulation framework conditioned on both the plant appearance and temporal growth stage, enabling realistic shade generation for unseen plants across their development.

Shadows encode rich information about scene geometry, illumination, and spatial relationships. Traditional shadow rendering relies on physics-based methods such as shadow mapping, shadow volumes, and ray tracing~\cite{williams1978casting,hu2024unveiling}, which require explicit 3D geometry and lighting parameters. In agriculture, Huang et al.~\cite{huang2024synthetic} generated a synthetic dataset of 50,000 images with physics-based shadow masks using 3D modeling software. However, it is limited to a resolution of 512$\times$512 and only a few plant models.

Learning-based shadow generation has evolved from GANs~\cite{liu2020arshadowgan,zhang2019shadowgan} to diffusion models~\cite{liu2024shadowdiffusion}. Existing methods focus on generic scene compositing and do not capture the dynamic, self-occluding nature of plant canopies under changing growth stages and lighting. Our approach addresses these challenges by conditioning on plant appearance and temporal growth stage, enabling realistic shade generation for unseen plants throughout development.
% \subsection{Generative Models for Simulation}

\section{PlantShade Dataset}
\subsection{Data Generation Framework Overview}

\begin{figure}[t]
    \centering
    \includegraphics[width=.95\linewidth]{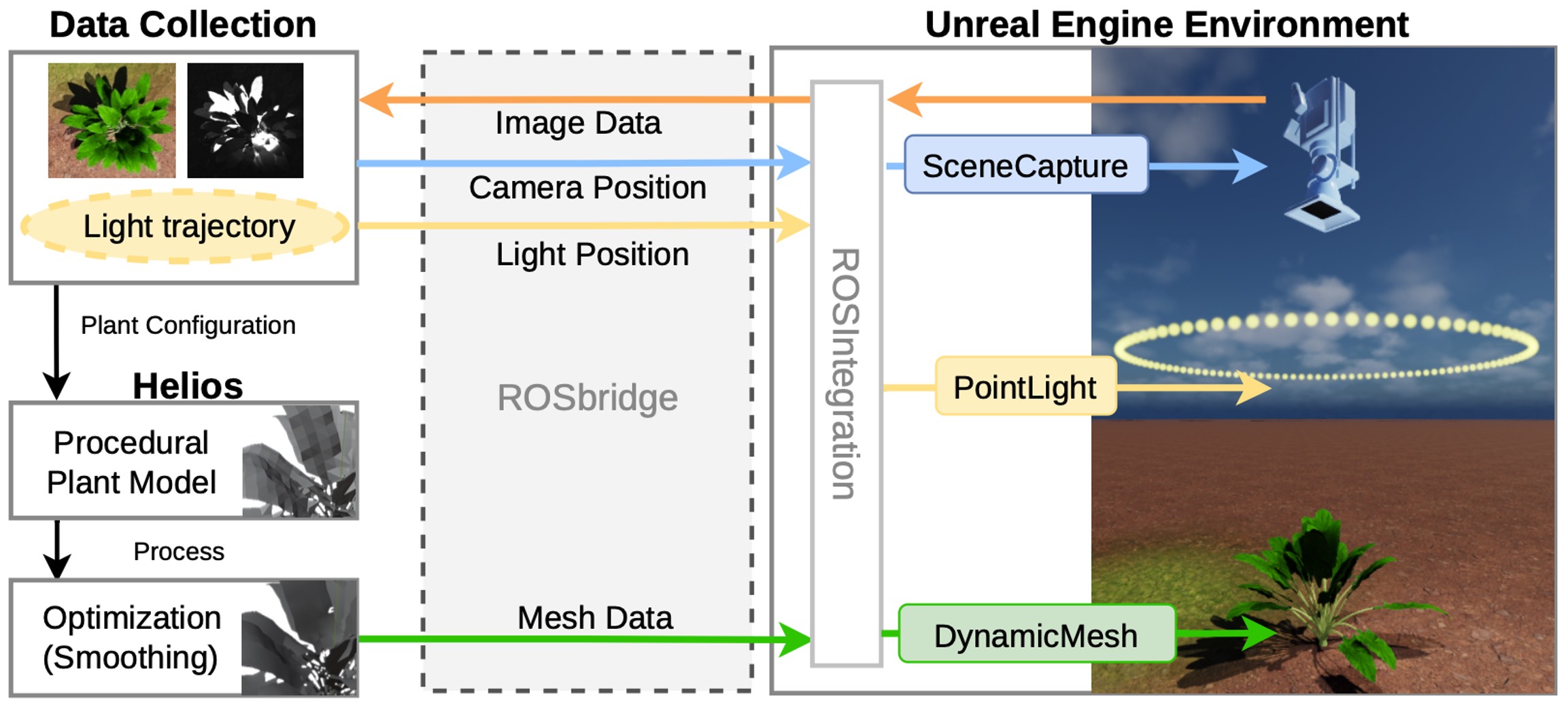}
    \includegraphics[width=.95\linewidth]{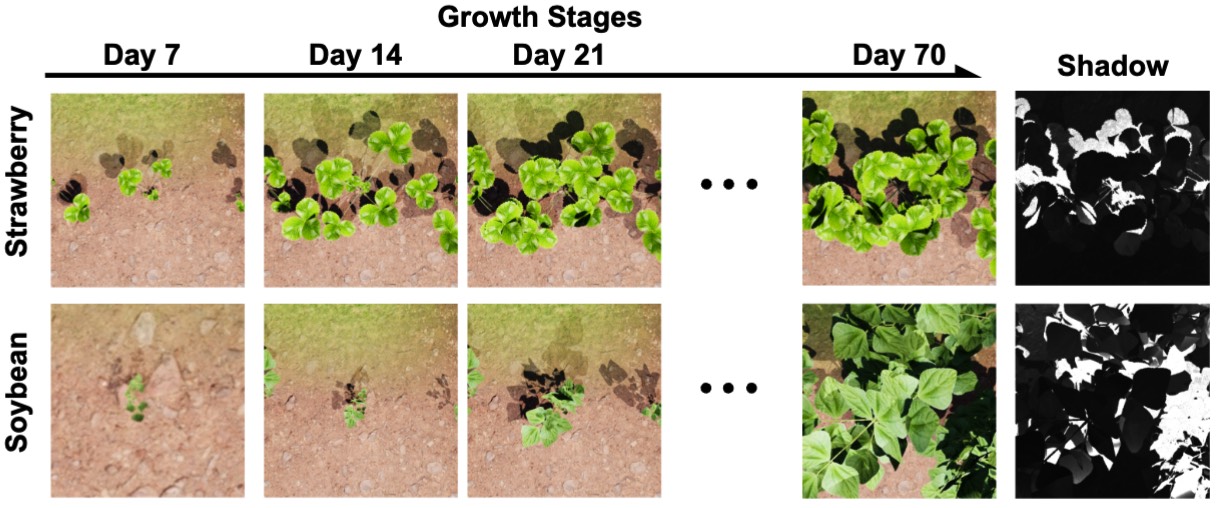}
    \caption{Simulation framework and data generation. Top: Data generation setup and flowchart with sugar beet in the sample. Bottom: Zoomed-in sample images for strawberry and soybean across growth stages.}
    \label{fig:simflowchart}
    \vspace{-2mm}
\end{figure}

To generate high-fidelity synthetic data with real-time performance, we developed a hybrid pipeline as shown in Fig.~\ref{fig:simflowchart} that integrates the biophysical procedural modeling capabilities of the Helios framework~\cite{bailey2019helios} with the photorealistic real-time rendering engine of Unreal Engine 5 (UE5) framework~\cite{sempnbv}. This architecture bridges the two simulation frameworks using ROSbridge %~\cite{rosbridge} 
and ROSIntegration~\cite{mania19scenarios} to enable mesh streaming in UE5. The mesh data is defined in the OBJ format and the file is sent to UE5 as string packets to be parsed and loaded using \textit{DynamicMeshComponent}, where the materials are provided in the MTL format that indicate the vertex color or a texture image. The data generation framework supports all 25 plant models from Helios with adjustable mesh parameters and age-based plant growth.
We utilize the UE5 framework to enable real-time smooth rendering of the RGB, plant mask, and shadow intensity ground truths. %Unlike Helios's radiation module that uses raytracing for light interception, 
The shadow intensity color $C_{shade}$ in UE5 is computed using a post processing material attached to the SceneCapture cameras, defined in Eq.~(\ref{eq:postprocessshade}). The luminance operator $D(\cdot)$, given in Eq.~(\ref{eq:postprocessdesaturate}), computes perceived luminance value from RGB values using standard sRGB weighting coefficients. This method produces an inverse luminance measure (shade) normalized by the material’s luminance, while the subtractive term enhances contrast in regions with intermediate shadow intensities.
\begin{equation}
C_{shade} = \frac{D(C_{D})}{D(C_{S})}-D(C_{D})
\label{eq:postprocessshade}
\end{equation}
\begin{equation}
D(C) = 0.213I_R+0.715I_G+0.072I_B
\label{eq:postprocessdesaturate}
\end{equation}
\noindent
 where $C_D$ denotes the material diffuse color (base color from the texture), $C_S$ is the final scene color, and $C=(I_R, I_G, I_B)$ represents the RGB channel values.

The number of leaf and stem subdivisions is significantly reduced from the default to accelerate the growth simulation and the mesh streaming process. The generated mesh also under goes an additional preprocessing step to enable smooth shading and optimized geometry order, preserving the natural appearances under the reduced geometry complexity.
\begin{table}[t]
\centering
\caption{Key specifications of the PlantShade dataset.}
\label{tab:dataset_specs}
\begin{tabular}{ll}
\hline
\textbf{Parameter} & \textbf{Specification} \\
\hline
Total Image Pairs & 38,500 \\
Resolution & $1920 \times 1080$ \\
Camera Height & 1.8 m (Nadir View) \\
Plant Species & 4 (\textit{Tomato}, \textit{Soybean}, \textit{Sugar beet}, \textit{Strawberry}) \\
Plant Age Range & 7 -- 119 days \\
Layout Configurations & $1\times1$, $1\times3$, $3\times5$ \\
Solar Elevation & 52.24$^\circ$ \\
Data Modalities & RGB, Plant Mask, Shadow Mask \\
\hline
\end{tabular}
\end{table}
\subsection{Dataset Overview and Statistics}

The PlantShade dataset consists of 38,500 synthetic RGB images at 1920$\times$1080 resolution, each paired with corresponding pixel-wise plant and shadow masks that indicate the plant region and the shadow intensity. Table~\ref{tab:dataset_specs} summarizes the distribution of the generated data. During dataset generation, the environment runs at an average of 100 frames per second (FPS), while the image data are published at 2 FPS. To account for lighting updates and data synchronization overheads, image pairs are saved at 6-second intervals to ensure matching image and supplementary light positions. %The dataset is divided into \textcolor{red}{X} training, \textcolor{red}{Y} validation, and \textcolor{red}{Z} testing pairs.

To ensure morphological diversity, we selected four crop models with distinct leaf and plant characteristics: Tomato (\textit{Solanum lycopersicum}), Soybean (\textit{Glycine max}), Sugar beet (\textit{Beta vulgaris}), and Strawberry (\textit{Fragaria $\times$ ananassa}). We varied the plant age between 7 and 119 days to capture a full range of growth stages. The scene layout was randomized using grid configurations of $1\times1$, $1\times3$, and $3\times5$ with 50~cm spacing, with growth intervals of either 1 or 7 days to simulate varying levels of canopy density and occlusion complexity. The data was collected from a nadir (top-down) perspective with the camera fixed at 1.8~m above the ground. Lighting environment includes a primary solar source with an elevation angle of 52.24$^{\circ}$ and a supplementary light source orbiting the plant at a height of 1.8~m with a 1.25~m radius.
\begin{figure}
    \centering
    \includegraphics[width=0.5\linewidth]{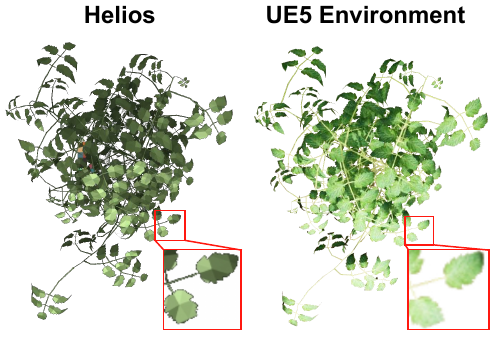}
    \includegraphics[width=0.46\linewidth]{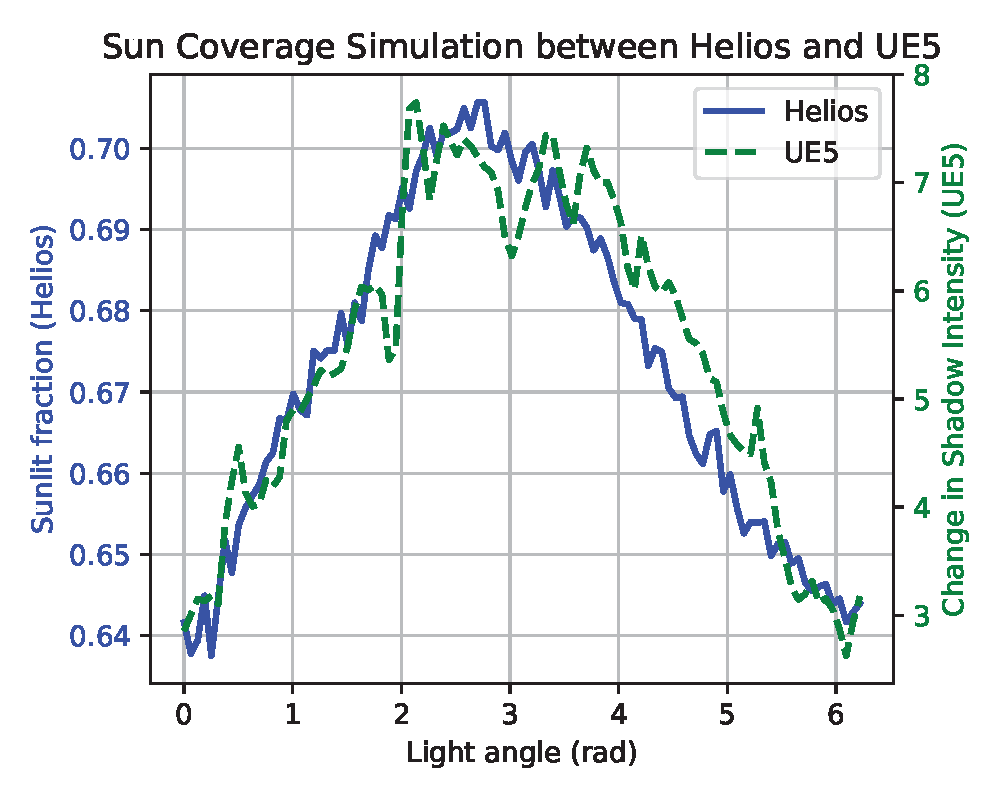}
    \caption{UE5 and Helios comparison for a sunlit tomato canopy. Left: Differences in visual appearance between renderings produced by Helios’s \textit{radiation} module and UE5, with the background removed for canopy shade clarity. Right: Sunlit canopy coverage under a rotating supplementary light source, demonstrating similar illumination trends between Helios (computed from all leaves) and UE5 (estimated from observed leaves only).}
    \label{fig:simcompare}
    \vspace{-3mm}
\end{figure}
\subsection{Comparison with Existing Approaches}
The synthetic shadow dataset generated by Huang et al.~\cite{huang2024synthetic} is limited to seven static plant models, whereas ours includes 385 models spanning four crop types and multiple growth stages. Existing 3D plant light intercept tools, such as Helios~\cite{bailey2019helios} and GroIMP~\cite{groimp} support diverse plant geometries and inter-plant resolution, and prioritize radiative-transfer accuracy rather than the photorealistic real-time rendering targeted in this work. Our pipeline is therefore complementary.

\begin{figure*}[t]
    \centering
    \includegraphics[width=0.99\linewidth]{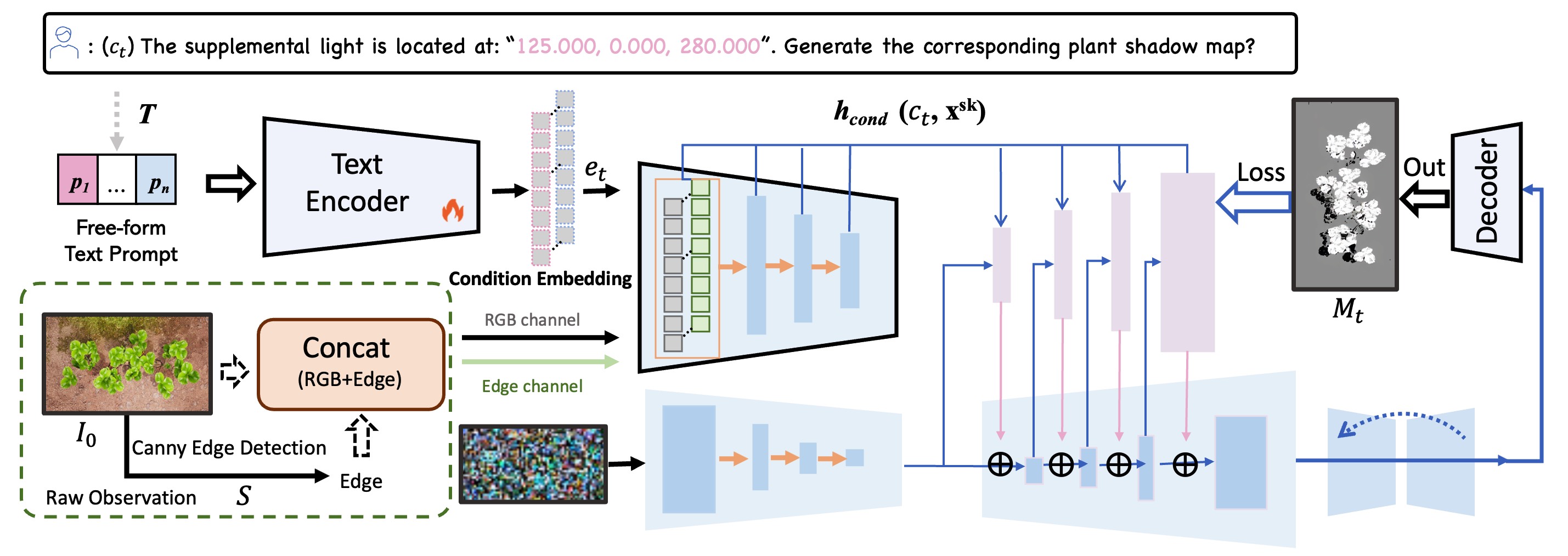}
    \caption{Architecture of the PlantShade shadow generative model. The framework combines a text-encoded supplementary light position with structural guidance from RGB and edge features to condition a diffusion backbone, generating shadow maps consistent with plant geometry and lighting configuration. }
    \label{fig:generativeshade}
    \vspace{-5mm}
\end{figure*}

%\subsubsection{Limitations of Helios}
Helios is a flexible, open-source software for 3D biophysical modeling with 25 procedural plant models, but its rendering modules (\textit{radiation} and \textit{visualizer} plugins) are primarily designed for biophysical visualization rather than photorealistic image generation. The \textit{radiation} plugin supports multiple light sources but lacks smooth shading, leading to visibly faceted geometry in rendered imagery. The OpenGL-based \textit{visualizer} plugin supports smooth shading, but is limited to a single light source. Fig.~\ref{fig:simcompare} provides a comparison of the same plant model computed in Helios's \textit{radiation} plugin and our framework, illustrating a more refined shade gradient and a similar shade curve across dynamic supplementary light positions on the same plant.

SynavisUE~\cite{baker2024scalable} integrates CPlantBox~\cite{zhou2020cplantbox} with UE5 for visualization and data generation. Their method computes the light interception via a material-based shader assigned to individual plants for light intensity. In contrast, our sensor-based shader attaches to the camera, enabling simultaneous rendering of brightness and surface normals without material switching in the plant models.

% \section{PlantShade Dataset\longchao{Xingjian}}

% \subsection{Dataset Overview and Statistics}
% Plant species and growth stages, Single-plant and multi-plant configurations

% \subsection{Plant Growth Model and Circular Multi-View Acquisition}

% \begin{itemize}
%     \item 360 Degree circular camera trajectory
%     \item 100 viewpoints per growth stage
%     \item RGB images, plant masks, and shadow maps
% \end{itemize}

% \paragraph{Comparison with Existing Datasets. A discussion}
% \begin{figure*}[t]
%     \centering
%     \includegraphics[width=0.99\linewidth]{figs/structure1.pdf}
%     \caption{The structure of the PlantShade's generative shade model.}
%     \label{fig:generativeshade}
% \end{figure*}
% \section{PlantShade Dataset\longchao{Xingjian}}

% \subsection{Dataset Overview and Statistics}
% Plant species and growth stages, Single-plant and multi-plant configurations

% \subsection{Plant Growth Model and Circular Multi-View Acquisition}

% \begin{itemize}
%     \item 360 Degree circular camera trajectory
%     \item 100 viewpoints per growth stage
%     \item RGB images, plant masks, and shadow maps
% \end{itemize}

% \paragraph{Comparison with Existing Datasets. A discussion}

% *Data set preparartion*

\section{Generative Plant Shade Simulation Model}

\subsection{Problem Formulation}

We formulate plant shade generation under supplementary lighting as a conditional generative modeling problem~\cite{zhang2023adding, da2025deepshade} shown in Fig.~\ref{fig:generativeshade}. 
We consider a fixed plant scene observed from a top-down camera view, where $\mathbf{I}_0 \in \mathbb{R}^{H \times W \times 3}$ denotes the base RGB image representing the plant appearance under a reference lighting configuration.
To preserve geometric and structural consistency across lighting conditions, we extract the representation $\mathbf{S} \in \mathbb{R}^{H \times W}$ (e.g., Canny edge) from $\mathbf{I}_0$.

We then introduce a complementary light source whose spatial position varies over time while the plant geometry and reference illumination remain fixed. 
At time step $t$, the complementary light configuration is described by a textual condition $c_t$, which parameterizes its location along a predefined circular trajectory. 
The complete conditioning input at time $t$ is therefore defined as 
$\mathcal{C}_t = \{\mathbf{I}_0, \mathbf{S}, c_t\}$. Our objective is to generate a time-dependent shadow map:
\begin{equation}
    \mathbf{M}_t \in \mathbb{R}^{H \times W}
\end{equation}
,where $\mathbf{M}_t$ is a continuous grayscale shadow field. It reflects the shadow occupancy induced by the complementary light configuration $c_t$. 
The sequence $\{\mathbf{M}_t\}_{t=1}^{T}$ captures the dynamics of plant shadows as the light moves.

\subsection{PlantShade Generative Simulation Model}

To realize the conditional mapping, we implemented a ControlNet-based diffusion architecture that generates $\mathbf{M}_t$ conditioned on $\mathcal{C}_t$, combining semantic light control with spatial structural guidance.

\paragraph{Light-Conditioned Structural Control.}
For each time step $t$, the textual condition $c_t$ describing the complementary light configuration is first encoded into a semantic embedding $\mathbf{e}_t$ using a pretrained text encoder. 
The embedding $\mathbf{e}_t$ is injected into the diffusion backbone via cross-attention layers, enabling the model to associate complementary light positions with corresponding shadow transformations. In parallel, to preserve geometric consistency across lighting variations, we introduce a ControlNet branch that receives the structural representation $\mathbf{S}$ together with the appearance prior $\mathbf{I}_0$. 
These spatial conditions are fused and injected into intermediate layers of the diffusion U-Net through learnable control connections, producing conditional feature maps that guide shadow generation. 

By jointly leveraging semantic light conditioning and structural control, the model ensures that variations in $\mathbf{M}_t$ are driven by changes in $c_t$ while maintaining consistent plant geometry across time steps.

\begin{table*}[t]
\centering
\small
\caption{Quantitative comparison between PlantShade and Stable Diffusion under different layout configurations. \cellcolor{blue}{Purple} columns marked with \textbf{*} indicate out-of-domain (OOD) settings.}
\label{tab:quantitative_results_combined}
\resizebox{0.99\textwidth}{!}{
\begin{tabular}{llcccc cccc cccc}
\toprule
& & \multicolumn{4}{c}{\textbf{Single}}
& \multicolumn{4}{c}{\textbf{$1\times3$}}
& \multicolumn{4}{c}{\textbf{$3\times5$}} \\
\cmidrule(lr){3-6}
\cmidrule(lr){7-10}
\cmidrule(lr){11-14}
\textbf{Metric} & \textbf{Method}
& Tomato & Soybean & Strawberry & \cellcolor{blue}{Sugar beet*}
& Tomato & Sugar beet & Soybean & Strawberry
& Tomato & Sugar beet & \cellcolor{blue}{Soybean*} & \cellcolor{blue}{Strawberry*} \\
\midrule

\multirow{2}{*}{mIoU $\uparrow$}
& PlantShade
& 0.2636 & 0.4371 & 0.3413 & \cellcolor{blue}{0.0314}
& 0.2561 & 0.4088 & 0.3574 & 0.3970
& 0.3283 & 0.5530 & \cellcolor{blue}{0.0863} & \cellcolor{blue}{0.1094} \\
& Diffusion
& 0.0080 & 0.0098 & 0.0033 & \cellcolor{blue}{0.0089}
& 0.0099 & 0.0082 & 0.0191 & 0.0095
& 0.0292 & 0.0349 & \cellcolor{blue}{0.0415} & \cellcolor{blue}{0.0441} \\
\midrule

\multirow{2}{*}{B-IoU $\uparrow$}
& PlantShade
& 0.3808 & 0.3921 & 0.3911 & \cellcolor{blue}{0.0451}
& 0.3661 & 0.3730 & 0.3390 & 0.3587
& 0.3996 & 0.4401 & \cellcolor{blue}{0.1238} & \cellcolor{blue}{0.1234} \\
& Diffusion
& 0.0176 & 0.0158 & 0.0073 & \cellcolor{blue}{0.0119}
& 0.0235 & 0.0143 & 0.0309 & 0.0162
& 0.0550 & 0.0481 & \cellcolor{blue}{0.0554} & \cellcolor{blue}{0.0564} \\
\midrule

\multirow{2}{*}{LPIPS $\downarrow$}
& PlantShade
& 0.0660 & 0.0638 & 0.0428 & \cellcolor{blue}{0.3713}
& 0.0925 & 0.0671 & 0.1251 & 0.0848
& 0.2936 & 0.2119 & \cellcolor{blue}{0.6464} & \cellcolor{blue}{0.6575} \\
& Diffusion
& 1.0366 & 0.9878 & 1.0713 & \cellcolor{blue}{0.9713}
& 1.0208 & 0.9713 & 1.0356 & 0.9321
& 0.9177 & 0.8123 & \cellcolor{blue}{0.8300} & \cellcolor{blue}{0.7895} \\
\midrule

\multirow{2}{*}{MSE $\downarrow$}
& PlantShade
& 72.157 & 50.832 & 64.363 & \cellcolor{blue}{74.116}
& 71.710 & 51.045 & 53.750 & 49.873
& 57.421 & 41.876 & \cellcolor{blue}{61.975} & \cellcolor{blue}{58.105} \\
& Diffusion
& 92.067 & 96.313 & 91.816 & \cellcolor{blue}{95.463}
& 93.596 & 93.891 & 95.513 & 95.392
& 94.773 & 94.572 & \cellcolor{blue}{97.970} & \cellcolor{blue}{97.381} \\
\bottomrule
\end{tabular}
}
\end{table*}
\paragraph{Conditional Diffusion and Light-Driven Shadow Evolution.}
During training, each shadow map $\mathbf{M}_t$ is progressively corrupted with Gaussian noise according to the forward diffusion process. 
At diffusion step $\tau$, the network predicts the noise residual
\begin{equation}
\epsilon_\theta(\mathbf{z}_\tau, \mathcal{C}_t, \tau)   
\end{equation}
where $\mathbf{z}_\tau$ denotes the noisy latent representation of $\mathbf{M}_t$. 
The model is optimized using the standard denoising objective
\begin{equation}
    \mathcal{L}_{\text{diff}} = 
\mathbb{E}_{t,\tau,\epsilon}
\left[
\|\epsilon - \epsilon_\theta(\mathbf{z}_\tau, \mathcal{C}_t, \tau)\|_2^2
\right]
\end{equation}

After iterative denoising, the latent representation is decoded to produce the generated shadow map $\mathbf{M}_t$.

Because $\mathbf{I}_0$ and $\mathbf{S}$ remain fixed across time steps, the learned model captures the complementary light–shadow interaction under varying $c_t$. 
As $c_t$ changes smoothly along the predefined trajectory, the generated sequence $\{\mathbf{M}_t\}_{t=1}^{T}$ exhibits continuous and physically coherent shadow evolution, enabling downstream evaluation of alternative light placements for improved plant illumination.

\begin{figure}
    \centering
    \includegraphics[width=1\linewidth]{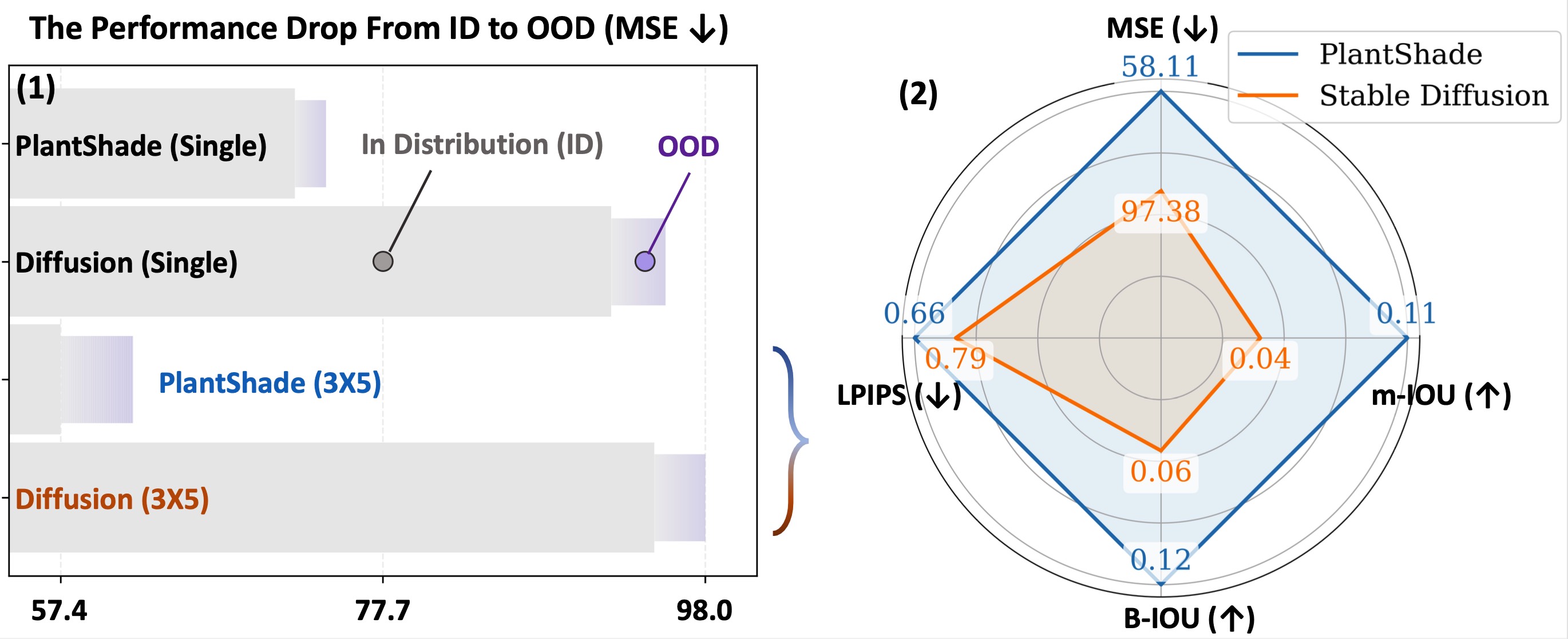}
   \caption{
Performance from in-distribution (ID) to OOD settings.
\textbf{(1)} MSE comparison under both \texttt{Single} and \texttt{$3\times5$ configurations}  shows that both PlantShade and Diffusion perform worse in OOD scenarios, indicating that OOD plant placement is consistently more challenging.
\textbf{(2)} However, the Radar comparison on the $3\times5$ OOD setting still demonstrates that PlantShade consistently outperforms the Diffusion baseline across four metrics, highlighting stronger robustness under distribution shift.
}
    \label{fig:ladarOOD}
    \vspace{-3mm}
\end{figure}

\section{Experiments}
\subsection{Experimental Setup}

\paragraph{Dataset}
% PlantShade contains 38,500 simulated light–shadow pairs across tomato, soybean, sugarbeet, and strawberry scenes, including both single-plant and multi-plant layouts, 7,600 pairs were kept out of the experiment dataset for \textbf{out-of-domain (OOD)} evaluation and 30,900 images were used for training and \textbf{in-domain (ID)} evaluations. 
PlantShade contains 38,500 simulated light–shadow pairs across tomato, soybean, sugarbeet, and strawberry scenes, including both single-plant and multi-plant layouts. Of these, 7,600 pairs were held out for out-of-domain (OOD) evaluation, and 30,900 were used for training and in-domain (ID) evaluation.
Each sample consists of an RGB observation, the corresponding shadow map, and a textual prompt describing the complementary light position. 
We randomly shuffle all samples (seed~=~42) and split them into 80\% training (24,720) and 20\% testing (6,180) sets at the sample level.

\paragraph{Training Details}
All models are trained for 50 epochs under identical settings. 
We adopt Stable Diffusion v2.1 as the diffusion backbone with an integrated ControlNet branch for structural conditioning. 

% \subsection{Quantitative Evaluation}

% All quantitative evaluations are conducted on the held-out test set. 
% We report results separately for different plant types (tomato, soybean, strawberry) and scene difficulty levels, including single-plant, $1\times3$, and $3\times5$ multi-plant layouts. 

% We evaluate shadow prediction accuracy using  MSE for structural similarity and intensity consistency, mean Intersection-over-Union (mIoU) for shadow region overlap, Boundary IoU (B-IoU) for contour alignment, and LPIPS for perceptual similarity. 
% Shadow regions are obtained by binarizing grayscale maps with a fixed threshold ($\tau=40$). 
%\subsection{Quantitative Evaluation}
\subsection{Evaluation Settings and Metrics}

All quantitative evaluations are conducted on the held-out test set and consist of two complementary settings: \textbf{ID} evaluation and \textbf{OOD} evaluation. Results are summarized in Table~\ref{tab:quantitative_results_combined}, with OOD comparisons further visualized in Figure~\ref{fig:ladarOOD}. 
For metrics, MSE is used to measure pixel-wise shadow prediction error. Mean Intersection-over-Union (mIoU) is used to quantify shadow region overlap, while Boundary IoU (B-IoU) measures contour alignment accuracy. LPIPS is adopted to assess perceptual similarity. Shadow regions are obtained by binarizing grayscale shadow maps with a fixed threshold ($\tau = 40$). Because plant shadows are thin and semi-transparent, IoU computed from binarized shadow masks is inherently conservative and should therefore be interpreted comparatively rather than in absolute terms. All metrics are reported as mean values over the respective test set splits.

\begin{figure*}
    \centering
    \includegraphics[width=.9\linewidth]{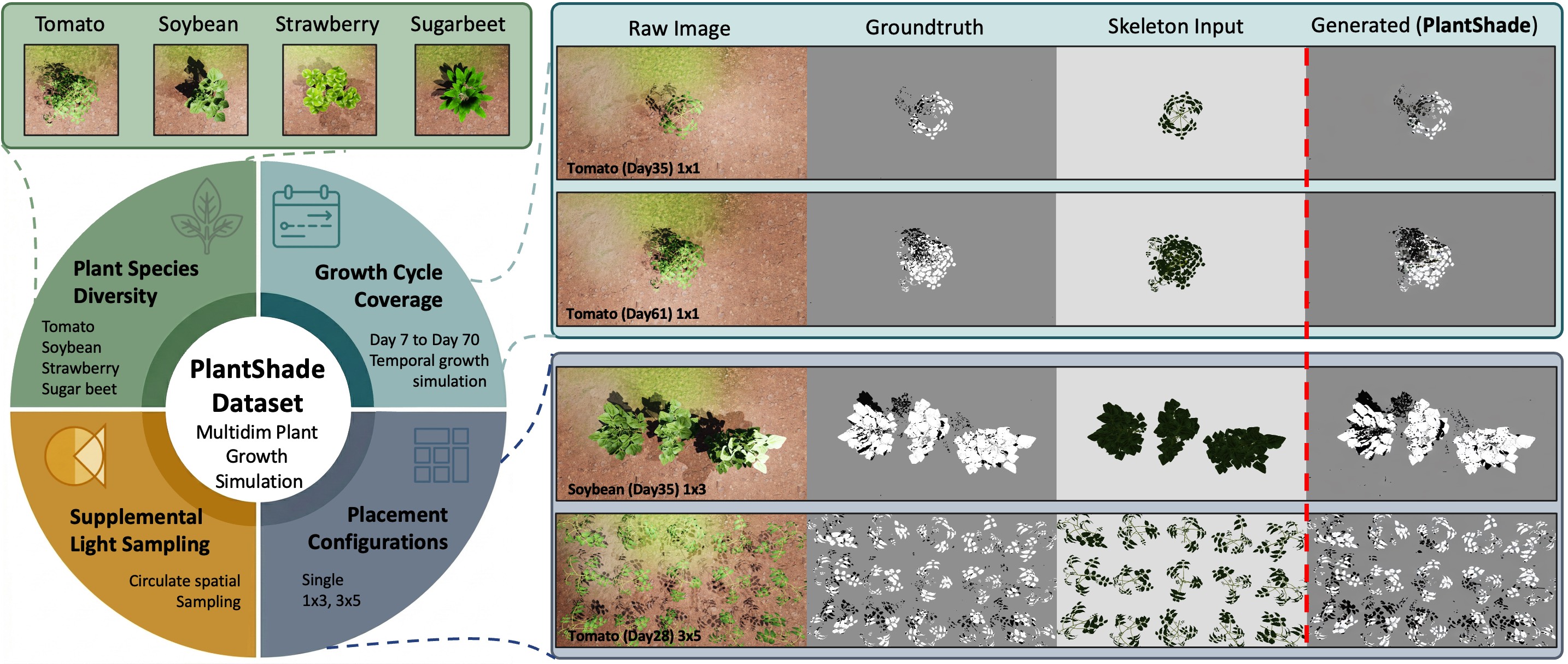}
   \caption{Overview of the PlantShade dataset and representative generation results. The left circular diagram summarizes the four core dimensions of the dataset: growth cycle coverage, plant species diversity, display configurations, and supplemental light sampling with 100 circular positions. The top-left panel shows representative samples of the four plant species included in the dataset. The top-right panel presents shadow estimation results across different growth stages, illustrating temporal consistency in light-driven shadow formation. The bottom-right panel shows generated shadow maps under varying spatial placement configurations, highlighting the model’s ability to generalize across density and layout complexity. The light sampling is shown in Fig.~\ref{fig:simflowchart}.
}
\vspace{-4mm}
    \label{fig:resultQualitive}
\end{figure*}

\subsection{Quantitative Evaluation}

Table~\ref{tab:quantitative_results_combined} reports quantitative comparisons between PlantShade and Stable Diffusion across plant species and layout complexities. In contrast to the previous version, we additionally include OOD configurations (highlighted in purple), enabling a direct assessment of generalization under unseen placement settings. Across all metrics (mIoU~$\uparrow$, B-IoU~\cite{da2025deepshade}~$\uparrow$, LPIPS~\cite{ghazanfari2023r}~$\downarrow$, and MSE~\cite{tan2013perceptually}~$\downarrow$), OOD scenarios consistently exhibit performance degradation compared to their in-distribution counterparts, indicating that denser or unseen plant-layout combinations introduce increased structural ambiguity and shadow overlap. Notably, this degradation is observed for both methods, confirming the intrinsic difficulty of distribution shift. Nevertheless, PlantShade maintains a clear margin over Stable Diffusion in both ID and OOD settings, demonstrating stronger robustness and more stable shadows under increasing layout complexity.

\subsection{Qualitative Evaluation}

We also conducted qualitative analysis across different plant species, growth stages, and spatial configurations as in Figure~\ref{fig:resultQualitive}. We evaluate shadow generation under multiple settings, including single-plant layouts, denser multi-plant arrangements ($1\times3$ and $3\times5$), and different temporal growth stages. We observe that PlantShade consistently captures \textbf{global light direction and structural shadow patterns} across species and placement layouts. Under single-plant configurations, the model preserves fine-grained structural details while producing coherent shadow regions. In denser multi-plant layouts, the model remains robust to increased occlusion and spatial interaction, although overlapping leaf structures introduce additional complexity. 
\textbf{Across growth stages} (e.g., early vs. later days), the generated shadows reflect the corresponding change in plant geometry and canopy density. The model adapts to increasing foliage complexity and maintains spatial alignment between plant structure and projected shadow, demonstrating temporal consistency in light-driven shadow simulation.
\textbf{Challenging scenarios:} The inspection reveals several challenging cases. Small plants are generally harder to model accurately, as their sparse and fine structures make shadow boundaries less distinct. In such cases, the model may over-generate shadow regions compared to the ground truth. Moreover, under dense $3\times5$ placements, complex inter-leaf occlusion leads to overlapping shadow regions, which increases ambiguity in shadow attribution across neighboring plants. Addressing these challenges is essential for improving shadow disentanglement and for supporting downstream photosynthesis modeling.

\subsection{Photosynthetic Gain Estimation}

Our PlantShade framework aims to maximize net photosynthesis by strategically placing supplemental lights to reduce the shaded area. For a given complementary light configuration, %$\mathbf{c}$, 
our model predicts the resulting shadow map. To quantify the downstream impact of shadow prediction on plant productivity, we translate the predicted shadow maps into estimates of leaf-level net photosynthesis using the non-rectangular hyperbola (NRH) model~\cite{thornley1990plantandcrop}.%,marshall1980C3model}.

\paragraph{NRH Model}
The net photosynthetic rate $P$ 
%($\mu$mol~CO$_2$~m$^{-2}$~s$^{-1}$) 
is modeled as a function of incident PAR $I$ as below $P(I)=$ 
\begin{equation}
 \frac{1}{2\theta}
\left[
\alpha I + P_{\max}
-
\sqrt{(\alpha I + P_{\max})^2 - 4\theta \,\alpha I\, P_{\max}}
\right]
\label{eq:nrh}
\end{equation}
where $\alpha$ is the apparent quantum yield (mol~CO$_2$~mol$^{-1}$~photons), $P_{\max}$ is the light-saturated maximum gross photosynthetic rate, and $\theta \in [0,1]$ is the convexity parameter governing the curvature of the transition from the light-limited to the light-saturated regime.

\paragraph{Leaf Area Computation from Depth Maps}

To convert pixel-level predictions into physically meaningful photosynthetic estimates, we compute the real-world leaf area using the depth maps rendered by the simulator. Under the pinhole camera model, a  pixel at depth $d_i$ subtends a small patch on the scene surface, and the total leaf area is obtained by summing over all leaf pixels $\mathcal{L}$:

\begin{equation}
A 
= \sum_{i \in \mathcal{L}} \frac{d_i^2}{f_x \, f_y},
\end{equation}
where $d_i$ is the depth value (mm), and $f_x = f_y = 872.72$ are the camera's focal lengths (pixels).

This formulation naturally accounts for perspective distortion, assigning smaller physical areas to closer surfaces. We note that this formulation assumes leaf surfaces are approximately fronto-parallel to the camera. For leaves with a significant tilt angle $\phi$ relative to the viewing direction, the true area would be $A_i/\cos\phi$; however, since our simulator renders top-down views of greenhouse canopies, this approximation is sufficient for the scenarios considered.

\paragraph{Photosynthesis Under Partial Shading} 
%Given a predicted shadow map, we partition each leaf's surface into a sunlit region and a shaded region. The shaded fraction is computed as an area-weighted ratio:
% \begin{equation}
% f = \frac{\sum_{i \in \mathcal{S}} A_i}{\sum_{i \in \mathcal{L}} A_i}
% \end{equation}
% where $\mathcal{S}\subset\mathcal{L}$ denotes the set of shaded leaf pixels. The effective canopy-level photosynthetic rate is then:
% \begin{equation}
% P_{\text{canopy}} = (1 - f)\cdot P(I_0) + f \cdot P(I_{\text{shadow}})
% \end{equation}
% where $I_0$ is the full irradiance on sunlit leaves and $I_{\text{shadow}}=\beta\cdot I_0$ is the attenuated irradiance in shaded regions, with $\beta$ denoting the light transmittance ratio.
Given a predicted shadow map, we compute the shaded fraction $f$ 
as the ratio of shaded leaf area to total leaf area:

\begin{equation}
f = \frac{\sum_{i \in \mathcal{S}} A_i}
         {\sum_{i \in \mathcal{L}} A_i}
\end{equation}
where $\mathcal{S} \subset \mathcal{L}$ denotes the set of shaded leaf pixels. 

Sunlit leaves receive full irradiance $I_0$, while shaded leaves receive 
attenuated irradiance $\beta \cdot I_0$, where $\beta$ is the light 
transmittance ratio. The canopy-level photosynthetic rate is then a weighted combination:

\begin{equation}
P_{\text{canopy}} 
= (1 - f)\, P(I_0) 
  + f\, P(\beta \cdot I_0)
\end{equation}

\paragraph{Photosynthetic Gain Results} 

We adopt the NRH parameterization from published physiological studies. Following~\cite{thornley1990plantandcrop}, the apparent quantum yield and convexity parameter are fixed at $\alpha = 0.05~\mathrm{mol~CO_2~mol^{-1}~photons}$ and $\theta = 0.7$ for all species, as these vary minimally across C$_3$ plants. 
The light-saturated photosynthetic rate $P_{\max}$ ($\mu\mathrm{mol~CO_2~m^{-2}~s^{-1}}$) is species-specific: $17.5$ for tomato~\cite{cannell1998tomato}, $32.5$ for soybean~\cite{slattery2017soybean}, $12.0$ for strawberry~\cite{hidaka2012strawberry}, and $20.0$ for sugar beet~\cite{tsialtas2012sugerbeet}. The ambient PAR and shading transmittance ratio are set to $I_0 = 1000~\mu\mathrm{mol~photons~m^{-2}~s^{-1}}$ and $\beta = 0.3$, representing typical greenhouse midday conditions.

\begin{figure}
    \centering
    \includegraphics[width=\linewidth]{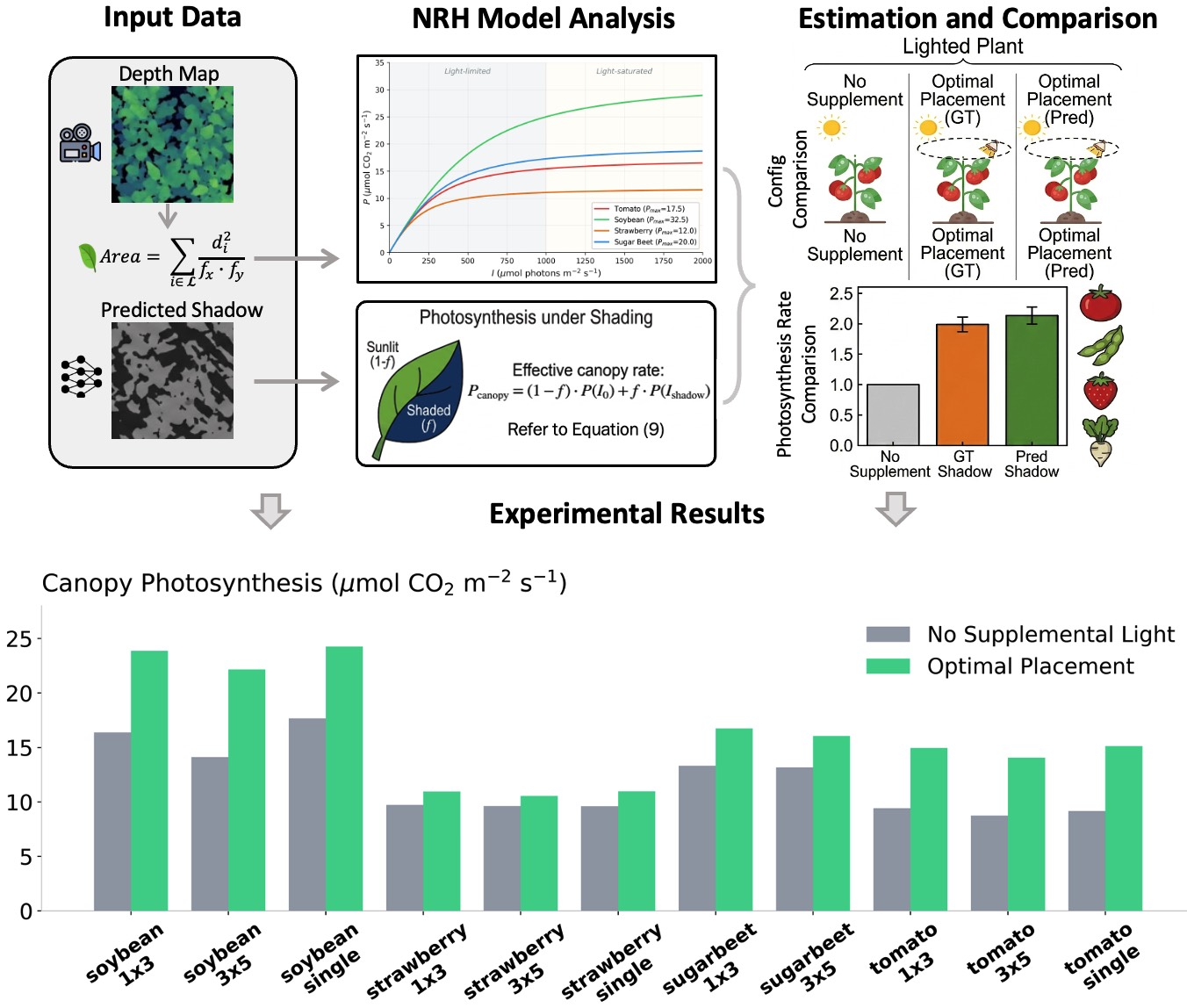}
    \caption{Overview of the photosynthetic gain estimation pipeline and results. Left: Depth and predicted shadow maps are used to estimate canopy geometry and shaded fraction. Middle: A species-specific light–response model converts shading into canopy photosynthetic rates. Right: Comparison of No Supplemental Light, Optimal Placement (GT), and Optimal Placement (Prediction). Bottom: Results for soybean, strawberry, sugar beet, and tomato under single, $1\times3$, and $3\times5$ layouts, showing consistent photosynthetic gains with optimal supplemental lighting.}
    \label{fig:photo_gain}
    \vspace{-8mm}
\end{figure}

Figure~\ref{fig:photo_gain} illustrates our photosynthetic gain estimation pipeline. Using rendered depth maps and predicted shadow maps, we first derive pixel-wise leaf area under the pinhole camera model and then apply the NRH model to estimate photosynthetic rates under sunlit and shaded conditions. The goal of this experiment is to demonstrate in-silico light-placement evaluation from predicted shadow maps rather than absolute photosynthetic gains.

Optimal supplemental light placement consistently enhanced canopy photosynthesis across all species and planting configurations. Soybean exhibited the largest absolute increase, with canopy photosynthesis rising from approximately 14--18 to 22--24 $\mu\mathrm{mol~CO_2~m^{-2}~s^{-1}}$. Sugar beet and tomato showed moderate improvements, whereas strawberry, which had the lowest $P_{\max}$, exhibited the smallest absolute gain. Overall, shadow-aware light placement appears to substantially improve canopy carbon assimilation, especially in species with high light-saturated photosynthetic capacity.

% We adopt species-specific NRH parameters from published physiological studies. For C3 plants, the apparent quantum yield $\alpha$ and convexity parameter $\theta$ have been shown to vary minimally across species~\cite{thornley1990plantandcrop}; we therefore fix $\alpha = 0.05~\mathrm{mol~CO_2~mol^{-1}~photons}$ and $\theta = 0.7$ for all four species in our experiments. 

% The light-saturated maximum photosynthetic rate $P_{\max}$, however, differs substantially among species. We survey reported values from the literature and adopted the following averaged estimates: 
% $P_{\max} = 17.5$ for tomato~\cite{cannell1998tomato}, 
% $32.5$ for soybean~\cite{slattery2017soybean}, 
% $12.0$ for strawberry~\cite{slattery2017strawberry}, and 
% $20.0~\mathrm{\mu mol~CO_2~m^{-2}~s^{-1}}$ for sugar beet~\cite{tsialtas2012sugerbeet}. 

% For all experiments, the ambient photosynthetically active radiation (PAR) is set to $I_0 = 1000~\mathrm{\mu mol~photons~m^{-2}~s^{-1}}$, and the light transmittance ratio in shaded regions is fixed at $\beta = 0.3$.
% These values represent typical greenhouse conditions during midday operations.

\section{Applications}

PlantShade's dataset and shade model enable several applications in agricultural robotics and crop management.

\paragraph{Supplemental lighting optimization and photosynthesis}
In controlled environments, supplemental lights extend photoperiod and improve yield. Our framework supports \emph{in silico} evaluation of light placement and trajectory: given a plant appearance and a candidate light configuration, the generative model predicts the resulting shadow map without full 3D rendering. We show that shadow maps can be combined with photosynthetic light-response models to estimate leaf-level or canopy-level carbon assimilation under different lighting scenarios. This allows growers and automated systems to compare alternative light layouts and schedules before physical deployment, reducing trial-and-error and energy waste.

\paragraph{Light exposure and microclimate analysis}
Spatially resolved shadow information is essential for understanding canopy light distribution and microclimate (e.g., temperature and humidity gradients that depend on shading). PlantShade's dynamic shadow sequences, conditioned on growth stage and light trajectory, provide a basis for analyzing how different crop layouts and light regimes affect light interception over time. Such analysis can inform planting density, row orientation, and the placement of sensors or robotic platforms that monitor crop health, supporting data-driven decisions in precision agriculture.

\paragraph{Robotic shade-aware pruning and canopy management}
Pruning strategies directly influence canopy architecture, direct energy to fruit production, and ultimately yield. By providing spatially resolved shadow maps under varying sun trajectories or supplemental lighting configurations, our framework enables quantitative evaluation of how individual branches contribute to self-shading within the canopy. For a given plant structure, simulated removal of candidate branches can be assessed \emph{in silico} by comparing resulting shadow distributions and predicted light interception. This analysis supports identification of pruning decisions that minimize excessive shading of lower or interior leaves, enhance canopy light uniformity, improve photosynthetic efficiency, and eventually increase yield potential.

% \paragraph{Synthetic data and sim-to-real for robot perception}
% Agricultural robots often operate under varying sun and supplemental lighting, so perception models must generalize across illumination conditions. The PlantShade dataset offers large-scale, pixel-aligned RGB and shadow pairs under controllable lighting, enabling robust training for tasks like segmentation, detection, or lighting-invariant feature learning. The generative model can further extend the data distribution to unseen species or growth stages, improving robustness and reducing the need for costly real-world data collection. Together, the dataset and model contribute to closing the sim-to-real gap for lighting-aware agricultural robotics.

\paragraph{Synthetic data and sim-to-real for robot perception}
Agricultural robots often operate under varying sun and supplemental lighting, so perception models must generalize across illumination conditions. The PlantShade dataset offers large-scale, pixel-aligned RGB and shadow pairs under controllable lighting, enabling robust training for tasks like segmentation, detection, or lighting-invariant feature learning. The generative model can further extend the data distribution to unseen species or growth stages, improving robustness and reducing the need for costly real-world data collection. 

\section{Conclusion}
We presented PlantShade, a real-time plant shade dataset and generative simulation framework for lighting-aware agricultural robotics. The dataset contains comprehensive image pairs across four crop species, multiple growth stages, and diverse spatial layouts under controlled circular supplementary lighting, explicitly modeling dynamic shadow evolution. We further proposed to leverage a conditional diffusion model that generates shadow maps based on plant appearance and light position (prompt). Quantitative and qualitative evaluations show that the PlantShade model consistently outperforms the baseline across metrics, while maintaining stronger robustness under out-of-distribution layout complexity. By linking predicted shadow maps to canopy-level photosynthetic estimation, our framework enables in silico evaluation of lighting strategies for robotic solutions in precision agriculture.

% \section*{APPENDIX}

% \section*{ACKNOWLEDGMENT}

% \tesay something

%%%%%%%%%%%%%%%%%%%%%%%%%%%%%%%%%%%%%%%%%%%%%%%%%%%%%%%%%%%%%%%%%%%%%%%%%%%%%%%%

% References are important to the reader; therefore, each citation must be complete and correct. If at all possible, references should be commonly available publications.

\bibliographystyle{IEEEtran}
\bibliography{software.bib}

\end{document}